\pdfoutput=1 
\documentclass[11pt]{article}
\usepackage[final]{coling}
\usepackage{times}
\usepackage{latexsym}
\usepackage[T1]{fontenc}
\usepackage[utf8]{inputenc}
\usepackage{amsmath}
\usepackage{amssymb}
\usepackage{microtype}
\usepackage{inconsolata}
\usepackage{graphicx}
\usepackage{float} 
\title{Enhancing Long-range Dependency with State Space Model and
Kolmogorov-Arnold Networks for Aspect-based Sentiment Analysis
}

\author{
	Adamu Lawan\textsuperscript{1}, Juhua Pu\textsuperscript{1,2}\thanks{Corresponding author}, Haruna Yunusa\textsuperscript{3}, Aliyu Umar\textsuperscript{4}, Muhammad Lawan\textsuperscript{5} \\
	\textsuperscript{1}School of Computer Science and Technology, Beihang University, Beijing, China \\
	\textsuperscript{2}National Research Institute for Teaching Materials in Information Science and Tech., China \\
	\textsuperscript{3}School of Automation Science and Electrical Engineering, Beihang University, Beijing, China \\
	\textsuperscript{4}School of Computing, University of Portsmouth, Portsmouth, UK \\
	\textsuperscript{5}Department of Information and Communication Technology, Federal University, Gusau, Nigeria \\
	\texttt{\{alawan, pujh, yunusa2k2\}@buaa.edu.cn}
}

\begin{document}
	\maketitle

	\begin{abstract}
	Aspect-based Sentiment Analysis (ABSA) evaluates sentiments toward specific aspects of entities within the text. However, attention mechanisms and neural network models struggle with syntactic constraints. The quadratic complexity of attention mechanisms also limits their adoption for capturing long-range dependencies between aspect and opinion words in ABSA. This complexity can lead to the misinterpretation of irrelevant contextual words, restricting their effectiveness to short-range dependencies. To address the above problem, we present a novel approach to enhance long-range dependencies between aspect and opinion words in ABSA (MambaForGCN). This approach incorporates syntax-based Graph Convolutional Network (SynGCN) and MambaFormer (Mamba-Transformer) modules to encode input with dependency relations and semantic information. The Multihead Attention (MHA) and Selective State Space model (Mamba) blocks in the MambaFormer module serve as channels to enhance the model with short and long-range dependencies between aspect and opinion words. We also introduce the Kolmogorov-Arnold Networks (KANs) gated fusion, an adaptive feature representation system that integrates SynGCN and MambaFormer and captures non-linear, complex dependencies. Experimental results on three benchmark datasets demonstrate MambaForGCN's effectiveness, outperforming state-of-the-art (SOTA) baseline models.
\end{abstract}

\section{Introduction}

In Natural Language Processing (NLP), text classification is vital for categorizing and extracting meaningful insights from textual data. A critical subset of text classification is sentiment analysis, which identifies the emotional tone or sentiment expressed within a text. With the growth of online platforms and the surge of user-generated content, sentiment analysis has become increasingly significant for applications like customer feedback analysis and product recommendation systems. However, traditional sentiment analysis often fails to capture sentiments about specific aspects or features within the text. This shortcoming led to the advent of Aspect-Based Sentiment Analysis (ABSA), which determines the overall sentiment and identifies and analyzes sentiments tied to particular aspects or features mentioned in the text. ABSA offers a more detailed and nuanced understanding of sentiment, providing valuable insights for businesses to enhance decision-making and improve user experiences.

Advancements in semantic-based models have significantly improved ABSA by combining various attention mechanisms. For instance, \citet{Tang2016} introduced a deep memory network that emphasizes the importance of individual context words by integrating neural attention models over external memory, effectively capturing complex sentiment expressions. Similarly, \citet{Wang2016} proposed an attention-based Long Short-Term Memory Network (LSTM) designed for ABSA, which uses attention mechanisms to highlight distinct sentence parts based on different aspects. Several researchers have developed interactive and multiple-attention mechanisms to enhance sentiment classification precision. Interactive Attention Networks (IAN), proposed by \citet{Ma2017}, facilitate interactive learning and generate distinct representations for targets and contexts. \citet{Peng2017} presented a framework utilizing a multiple-attention mechanism to capture sentiment features, integrating these attentions with a recurrent neural network for improved expressiveness. Similarly, \citet{Fan2018} introduced a multi-grained attention network (MGAN) that employs fine-grained attention mechanisms to capture word-level interactions between aspects and context. \citet{Wang2021} proposed a model using BERT for word embeddings, integrating intra-level and inter-level attention mechanisms and a feature-focus attention mechanism to enhance sentiment identification. Other studies have focused on integrating syntactic information and explicit knowledge into attention mechanisms. \citet{He2018} worked on better integrating syntactic information to capture the relationship between aspect terms and context. \citet{Ma2018} proposed augmenting LSTM with a stacked attention mechanism for target and sentence levels, introducing Sentic-LSTM to integrate explicit and implicit knowledge. Co-attention mechanisms have also been explored to enhance sentiment classification. \citet{Yang2019} introduced a co-attention mechanism alternating between target-level and context-level attention, proposing Coattention-LSTM and Coattention-MemNet networks. \citet{Cheng2022} advanced the field by presenting a multi-head co-attention network model with three modules: extended context, component focusing, and multi-headed co-attention, enhancing transformer-based sentiment analysis by improving context handling and refining attention mechanisms for multi-word targets.

In contrast, syntax-based models \citep{Sun2019, Zhang2019, Liang2022, Gu2023a, Wu2023, Liu2023, Li2023, Zhu2024} leverage syntactic information and word dependencies to improve ABSA. \citet{Sun2019} and \citet{Zhang2019} layered a GCN to extract comprehensive representations from the dependency tree. \citet{Liang2022, Wu2023, Gu2023a} integrated contextual knowledge into the GCN to improve ABSA.

Attention mechanisms in neural networks face notable challenges when addressing syntactic constraints, particularly in ABSA. Additionally, the quadratic complexity of standard attention mechanisms limits their ability to effectively capture long-range dependencies between aspect and opinion words. This limitation often leads to the misinterpretation of irrelevant contextual words, restricting the model's effectiveness to short-range dependencies. While some studies attempt to merge semantic and syntactic approaches, they often fall short in effectively integrating these two types of information, leading to suboptimal performance. Furthermore, a significant challenge in ABSA is handling implied opinion words—those that are not explicitly stated but still contribute to sentiment analysis. These implicit opinions can complicate aspect sentiment prediction, as traditional models rely heavily on explicit aspect-opinion pairs.

To address these challenges, we propose MambaForGCN, a novel framework specifically designed to enhance long-range dependency modeling in ABSA. The framework introduces a syntax-based SynGCN module, which encodes dependency relations to capture syntactic structures effectively. Complementing this, our innovative MambaFormer module enriches the model with semantic information through a combination of Multi-Head Attention (MHA) and Mamba blocks, enabling precise modeling of both short- and long-range dependencies between aspect and opinion words. This approach ensures that neither short-range nor long-range dependency constraints limit the framework’s ability to capture relevant contextual information.

Moreover, our use of Kolmogorov-Arnold Networks (KAN) gated fusion sets this framework apart. The gated fusion mechanism adaptively integrates feature representations from the SynGCN and MambaFormer modules, selectively filtering critical information for the ABSA task. By leveraging the non-linear dependency modeling capability of KANs, our framework can identify and infer sentiment even when opinion words are implied. KANs learn complex, non-linear relationships between words and aspects, detecting subtle sentiment patterns that may not be immediately obvious in the text’s surface structure. This ability makes KANs an ideal tool for capturing the nuanced, contextual sentiment expressed through implicit opinions, ultimately enhancing the robustness and accuracy of sentiment prediction.

The main contributions of this paper are as follows:
\begin{itemize}
	\item To the best of our knowledge, we introduce the selective state space model into ABSA for the first time, significantly enhancing the model’s ability to capture long-range dependencies.
	\item We leverage KANs to capture complex dependencies within the text. This novel application in ABSA enables MambaForGCN to identify and classify sentiment even when opinion words are implied.
	\item The experimental results on three benchmark datasets showcase the effectiveness of the MambaForGCN model, surpassing some state-of-the-art (SOTA) baselines.
\end{itemize}

\section{Related Work}

\citet{Tang2016} introduced a deep memory network for aspect-level sentiment classification, emphasizing the importance of individual context words by integrating neural attention models over external memory to capture sentiment nuances effectively. \citet{Ma2017} proposed Interactive Attention Networks (IAN) to facilitate interactive learning and generate distinct representations for targets and contexts, enhancing sentiment classification precision. \citet{Fan2018} introduced a multi-grained attention network (MGAN) that used fine-grained attention mechanisms to capture word-level interactions between aspects and context, enhancing classification accuracy. \citet{He2018} refined target representation and integrated syntactic information into the attention mechanism to better capture the relationship between aspect terms and context. \citet{Yang2019} introduced an attention mechanism alternating between target-level and context-level attention to improve sentiment classification. Cheng et al. (2022) proposed a component focusing on a multi-head co-attention network model, enhancing bidirectional encoder representations and improving the weighting of adjectives and adverbs.

\citet{Sun2019} proposed merging convolution over a dependency tree (CDT) with bi-directional long short-term memory (Bi-LSTM) to analyze sentence structures effectively. \citet{Liang2022} proposed Sentic GCN, a graph convolutional network based on SenticNet, to leverage affective dependencies specific to aspects. By integrating affective knowledge from SenticNet, the enhanced dependency graphs considered both the dependencies of contextual and aspect words and the affective information between opinion words and aspects. \citet{Wu2023} introduced KDGN, a knowledge-aware Dependency Graph Network that integrates domain knowledge, dependency labels, and syntax paths into the dependency graph framework, enhancing sentiment polarity prediction in ABSA tasks. \citet{Zhu2024} introduced a deformable convolutional network model that leverages phrases for improved sentiment analysis, using deformable convolutions with adaptive receptive fields to capture phrase representations across various contextual distances. The model also integrated a cross-correlation attention mechanism to capture interdependencies between phrases and words.

A notable area of research involves combining the Transformer and Mamba for language modeling \citep{Fathi2023, Lieber2024, Park2024, Xu2024}. Comparative studies have shown that Mambaformer is effective in in-context learning tasks. Jamba \citep{Lieber2024}, the first production-grade hybrid model of attention mechanisms and SSMs, features 12 billion active and 52 billion available parameters, demonstrating strong performance for long-context data. We are interested in using Mamba to capture long-term dependency in ABSA.

\section{Preliminaries}

\subsection{State Space Models (SSM)}

SSM-based models \citep{Gu2020, Gu2021, Gu2023} are based on continuous systems that map a 1-D input sequence $x(t)$ to an output sequence $y(t)$ via a hidden state $h(t)$. This system is defined using parameters $A \in \mathbb{R}^{N \times N}$, $B \in \mathbb{R}^{N \times 1}$, and $C \in \mathbb{R}^{1 \times N}$ as follows:
\begin{align}
	h'(t) &= A h(t) + B x(t) \tag{1} \\
	y(t) &= C h(t) \tag{2}
\end{align}

S4 and Mamba are discrete adaptations of this continuous system, utilizing a timescale parameter $\Delta$ to convert the constant parameters $A$ and $B$ into their discrete equivalents $\bar{A}, \bar{B}$ through a zero-order hold (ZOH) transformation:
\begin{align}
	\bar{A} &= \exp(\Delta A) \tag{3} \\
	\bar{B} &= (\Delta A)^{-1} (\exp(\Delta A) - I) \cdot \Delta A \tag{4}
\end{align}

The discrete form of the system, with step size $\Delta$, is given by:
\begin{align}
	h_t &= \bar{A} h_{t-1} + \bar{B} x_t \tag{5} \\
	y_t &= C h_t \tag{6}
\end{align}

Finally, these models compute the output using a global convolution:
\begin{align}
	\bar{K} &= (C \bar{B}, C \bar{A} \bar{B}, \ldots, C \bar{A}^{M-1} \bar{B}) \tag{7} \\
	y &= x \ast K \tag{8}
\end{align}
where $M$ represents the length of the input sequence $x$, and $\bar{K} \in \mathbb{R}^M$ is a structured convolutional kernel.

\subsection{Kolmogorov-Arnold Networks (KANs)}

KANs \citep{Liu2024} feature a distinctive architecture that differentiates them from traditional Multi-Layer Perceptrons (MLPs). Instead of using fixed activation functions at nodes, KANs employ learnable activation functions on the network edges. This fundamental change involves substituting conventional linear weight matrices with adaptive spline functions. These spline functions are parameterized and optimized during training, enabling a more flexible and responsive model architecture that dynamically adapts to complex data patterns.

The Kolmogorov-Arnold representation theorem asserts that a multivariate function $f(x_1, x_2, \ldots, x_n)$ can be represented as:
\begin{align}
	f(x_1, x_2, \ldots, x_n) = \sum_{q=1}^{2n+1} \Phi_q \left( \sum_{p=1}^n \varphi_{q,p}(x_p) \right) \tag{9}
\end{align}
In this context, $\varphi_{q,p}$ are univariate functions mapping each input variable $x_p$ as $\varphi_{q,p} : [0,1] \to \mathbb{R}$, and $\Phi_q : \mathbb{R} \to \mathbb{R}$ are also univariate functions.

KANs organize each layer into a matrix of these learnable 1D functions:
\begin{align}
	\Phi &= \{\varphi_{q,p}\} \tag{10} \\
	p &= 1, 2, \ldots, n_{\text{in}} \tag{11} \\
	q &= 1, 2, \ldots, n_{\text{out}} \tag{12}
\end{align}

Each function $\varphi_{q,p}$ is defined as a B-spline, a spline function created from a linear combination of basis splines, which enhances the network's ability to learn complex data representations. In this context, $n_{\text{in}}$ is the number of input features for a given layer and $n_{\text{out}}$ indicates the number of output features that the layer generates. The activation functions $\varphi_{l,i,j}$ within this matrix are implemented as trainable spline functions, formulated as:
\begin{align}
	\text{spline}(x) = \sum_i c_i B_i(x) \tag{13}
\end{align}
This formulation enables each $\varphi_{l,i,j}$ to adjust its shape according to the data, providing unparalleled flexibility in how the network captures input interactions.

The overall architecture of a KAN resembles stacking layers in MLPs, but it goes further by employing complex functional mappings instead of fundamental linear transformations and nonlinear activations:
\begin{align}
	\text{KAN}(x) = (\Phi_{L-1} \circ \Phi_{L-2} \circ \ldots \circ \Phi_0)(x) \tag{14}
\end{align}

\section{Proposed MambaForGCN Model}

Figure 1 gives an overview of MambaForGCN. In this section, we describe the MambaForGCN model, which is mainly composed of four components: the input and embedding module, the syntax-based GCN module, the MambaFormer module, and the KAN-gated fusion module. Next, components of MambaForGCN will be introduced separately in the rest of the sections.

\begin{figure*}[t] 
	\centering
	\includegraphics[width=\textwidth]{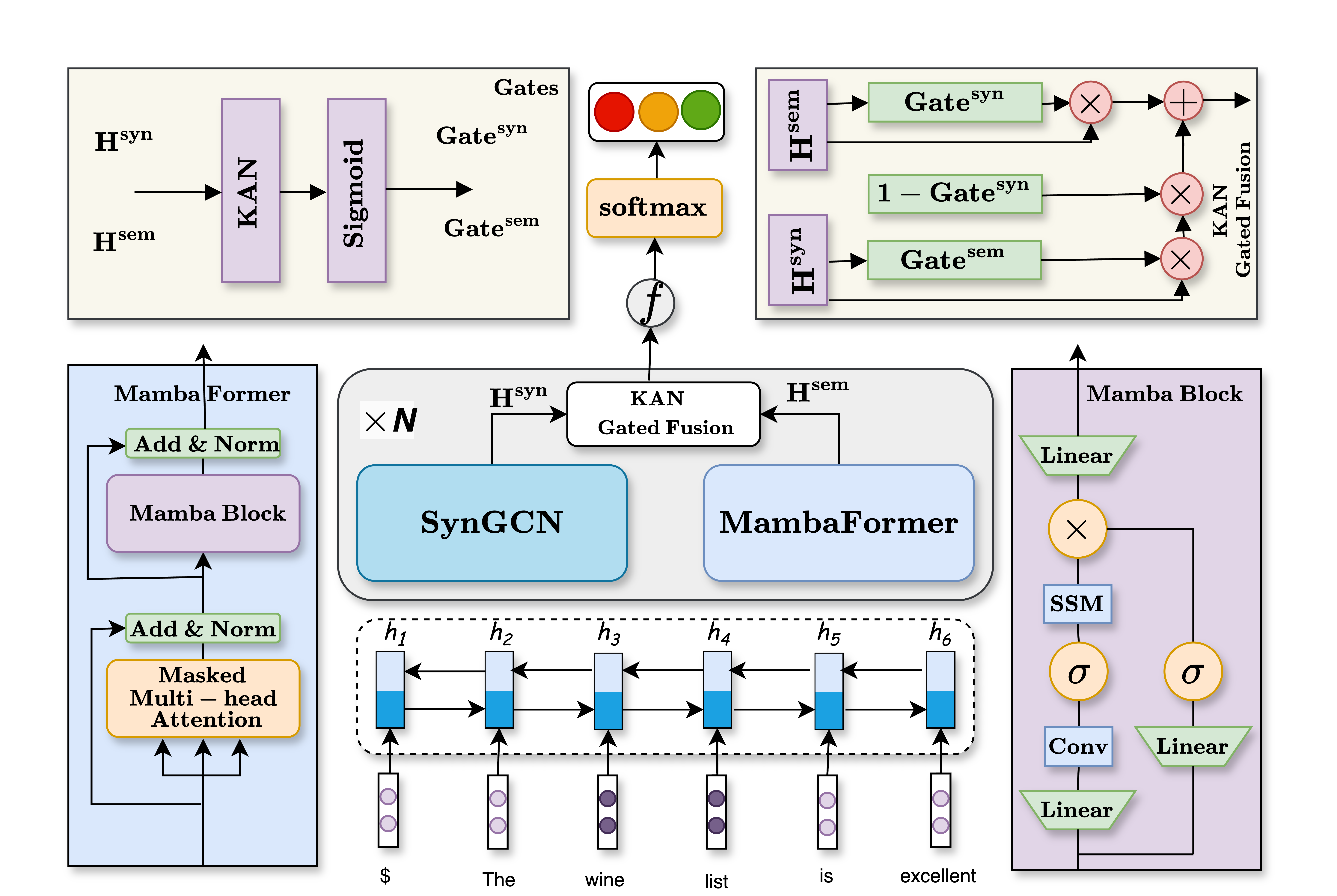}
	\caption{MambaForGCN complete architecture}
	\label{fig:MambaForGCN}
\end{figure*}

\subsection{Embedding Module}

Given a sentence $s$ and an aspect $a$ as a subset of $s$, we use a BiLSTM or BERT model for sentence encoding. Each word in $s$ is converted into a low-dimensional vector using an embedding matrix $E$, resulting in word embeddings $x$. These embeddings are fed into a BiLSTM to generate hidden state vectors $h_i$, capturing contextual information. The subsequence $h_a$, corresponding to the aspect term, is extracted from the hidden state matrix $H$ and used as the initial node representation in the MambaForGCN. For BERT, the input is formatted as "[CLS] sentence [SEP] aspect [SEP]," allowing BERT to capture complex relationships between opinion words and the aspect through its contextual embeddings.

\subsection{Syntax-based GCN Module}

The SynGCN module utilizes syntactic information as its input. Instead of relying on the final discrete output from a traditional dependency parser, we encode syntactic information using a probability matrix that represents all possible dependency arcs like in \cite{Li2021}. This method captures a broader range of syntactic structures, providing a more detailed and flexible understanding of sentence syntax. By considering the likelihood of multiple dependency arcs, this approach reduces the impact of potential errors in dependency parsing. We utilize the \cite{Mrini2019}, a cutting-edge model in the field of dependency parsing, to generate this probability matrix. The LAL-Parser's output is a probability distribution over all possible dependency arcs, effectively encapsulating the latent syntactic relationships within a sentence.

This comprehensive syntactic encoding allows SynGCN to understand complex grammatical structures. The SynGCN module uses a syntactic adjacency matrix $A^{\text{syn}} \in \mathbb{R}^{n\times n}$ to process the hidden state vectors $H$ from the BiLSTM, which act as the initial node representations in the syntactic graph. Through the SynGCN module, the syntactic graph representation $H^{\text{syn}}= \{ h_1^{\text{syn}}, h_2^{\text{syn}}, \ldots, h_n^{\text{syn}} \}$ is derived using equation (15). In this context, $h_i^{\text{syn}} \in \mathbb{R}^d$ represents the hidden state of the $i^{\text{th}}$ node.
\begin{equation}
	h_i^l = \sigma\left(\sum_{j=1}^n A_{ij} W^l h_j^{(l-1)} + b^l\right)\tag{15}
\end{equation}

\subsection{MambaFormer Module}

This module consists of two blocks, MHA and Mamba. They extract textual semantic information related to the given sentence and aspect. The MHA block captures short-range dependencies between aspect and opinion words, while the Mamba block captures long-range dependencies.

\textbf{Multihead Attention Block:} To extract important textual semantic information related to the given sentence and aspect, specifically for short-range word dependencies, we employ the MHA mechanism as shown in Fig. 1. In the MHA block, computation adheres to the standard process of transformer architecture \cite{Vaswani2017}. The first step in computing attention weights score is to take the dot product of the keys $K$ and queries $Q$. Next, another dot product between the score and the values $V$ yields the output representation $H^{\text{mha}'}$ of the attention module. The output \( H^{\text{mha}} \) represents a residual connection followed by layer normalization, which stabilizes and improves the training of the model by combining the output \( H^{\text{mha}'} \) with the original input \( h \), ensuring better gradient flow and normalized feature scaling. Below is an outline of this method:
\begin{align}
	K, Q, V &= h_j^{(l-1)} W_k, h_j^{(l-1)} W_q, h_j^{(l-1)} W_v\tag{16} \\
	\text{score} &= \frac{\text{softmax}(QK + \text{Mask})}{\sqrt{d_k}}\tag{17} \\
	H^{\text{mha}'} &= \text{score} \cdot V\tag{18} \\
	H^{\text{mha}} &= \text{LayerNorm}(H^{\text{mha}'} + h)\tag{19}
\end{align}

\textbf{Mamba Block:} Although transformers have proven effective in capturing dependency, their quadratic complexity of attention mechanism prevents their further adoption in long-range word dependencies, thus limiting them to the short-range range. To tackle this problem, we utilize the Mamba block. This approach ensures that essential connections and long-range dependencies between aspect word features and semantic emotional features are maintained throughout the analysis. As seen in Fig. 1, the Mamba block is designed for sequence-to-sequence tasks with consistent input and output dimensions. It expands the input $H^{\text{mha}}$ through two linear projections. One projection involves a convolutional layer and SiLU activation before passing through an SSM module, which filters irrelevant information and selects input-dependent knowledge. Simultaneously, another projection path with SiLU activation serves as a residual connection, combining its output with the SSM module's result via a multiplicative gate. Ultimately, the Mamba block outputs $H^{\text{mam}}$ in $H^{\text{mam}} \in \mathbb{R}^{B\times L\times D}$ through a final linear projection, providing enhanced sequence processing capabilities. Finally, $H^{\text{sem}}$ represents the output of the MambaFormer module after applying layer normalization to the sum of the outputs from the Mamba and MHA layers.
\begin{align}
	H^{\text{mam1}} &= \text{SiLU}(\text{Conv1D}(\text{Linear}(H^{\text{mha}})))\tag{20} \\
	H^{\text{mam2}} &= \text{SiLU}(\text{Linear}(H^{\text{mha}}))\tag{21} \\
	H^{\text{mam3}} &= \text{Linear}(\text{SSM}(H^{\text{mam1}}) \circ H^{\text{mam2}})\tag{22} \\
	H^{\text{mam}} &= \text{Linear}(H^{\text{mam3}})\tag{23} \\
	H^{\text{sem}} &= \text{LayerNorm}(H^{\text{mam}} + H^{\text{mha}})\tag{24}
\end{align}

\subsection{KAN Gated Fusion Module}

Gated fusion has demonstrated effectiveness in language modelling tasks \cite{Lawan2024, Zhao2024}. To dynamically assimilate valuable insights from the syntax-based GCN and MambaFormer, we used a KAN-gated fusion module to reduce interference from unrelated data. Gating is a potent mechanism for assessing the utility of feature representations and integrating information aggregation accordingly. This module uses a simple addition-based fusion mechanism to achieve gating, which controls the flow of information through gate maps, as shown in Fig. 1. Specifically, the representations $H^{\text{syn}}$ and $H^{\text{sem}}$ are associated with gate maps $\text{Gate}^{\text{syn}}$ and $\text{Gate}^{\text{sem}}$, respectively. These gate maps originate from a KANs using a one-dimensional layer. These gate maps are used to provide technical specifications for the gated fusion process:
\begin{align}
    \text{Gate}^{\text{syn}} &= \sigma(\text{KAN}(H^{\text{syn}}))\tag{25} \\
    \text{Gate}^{\text{sem}} &= \sigma(\text{KAN}(H^{\text{sem}}))\tag{26} \\
    H^{\text{c}} &= \text{Gate}^{\text{syn}} H^{\text{syn}} 
    + (1 - \text{Gate}^{\text{syn}}) \notag \\
    &\quad \times \text{Gate}^{\text{sem}} H^{\text{sem}}\tag{27}
\end{align}

We utilize mean pooling to condense contextualized embeddings $H^{\text{c}}$, which assists downstream classification tasks. Following this, we apply a linear classifier to generate logits. Finally, softmax transformation converts logits into probabilities, facilitating ABSA. Each component is essential in analyzing input text for ABSA tasks from the embedding layer to the sentiment classification layer.
\begin{align}
	H^{\text{mp}} &= \text{MeanPooling}(H^{\text{c}})\tag{28} \\
	p(a) &= \text{softmax}(W_p H^{\text{mp}} + b_p)\tag{29}
\end{align}

\subsection{Training}

We utilize the standard cross-entropy loss as our primary objective function:
\begin{equation}
	L(\theta) = -\sum_{(s,a) \in D} \sum_{c \in C} \log p(a)\tag{30}
\end{equation}
Computed over all sentence-aspect pairs in the dataset $D$. For each pair $(s,a)$, representing a sentence $(s)$ with aspect $(a)$, we compute the negative log-likelihood of the predicted sentiment polarity $p(a)$. Here, $\theta$ encompasses all trainable parameters and $C$ denotes the collection of sentiment polarities.

\section{Experiment}

\subsection{Datasets}

\begin{table}[h!]
	\centering
	\begin{tabular}{l l c c c}
		\hline
		\textbf{Dataset} & \textbf{Division} & \textbf{Pos} & \textbf{Neg} & \textbf{Neu} \\\hline
		Rest14 & Train & 2164 & 807 & 637 \\
		& Test  & 727  & 196 & 196 \\\hline
		Laptop14 & Train & 976  & 851 & 455 \\
		& Test  & 337  & 128 & 167 \\\hline
		Twitter & Train & 1507 & 1528 & 3016 \\
		& Test  & 172  & 169 & 336 \\\hline
	\end{tabular}
	\caption{Statistics of three benchmark datasets}
	\label{tab:dataset_statistics}
\end{table}

Table \ref{tab:dataset_statistics} provides comprehensive statistics for these datasets. Three publicly available sentiment analysis datasets are used in our experiments: the Twitter, the Laptop, and Restaurant 14 review datasets from the SemEval 2014 Task \citep{Pontiki2014}.

\subsection{Implementation}

The LAL-Parser \citep{Mrini2019} is used for dependency parsing, with word embeddings initialized by 300-dimensional pre-trained Glove vectors \citep{Pennington2014}.

\begin{table*}[t] 
	\centering
	\resizebox{\textwidth}{!}{
		\begin{tabular}{l cc cc cc}
			\hline
			\textbf{Model} & \multicolumn{2}{c}{\textbf{Restaurant14}} & \multicolumn{2}{c}{\textbf{Laptop14}} & \multicolumn{2}{c}{\textbf{Twitter}} \\
			& \textbf{Acc.} & \textbf{F1} & \textbf{Acc.} & \textbf{F1} & \textbf{Acc.} & \textbf{F1} \\\hline
			ATAE-LSTM \cite{Wang2016} & 77.20 & - & 68.70 & - & - & - \\
			IAN \cite{Ma2017} & 78.60 & - & 72.10 & - & - & - \\
			RAM \cite{Peng2017} & 80.23 & 70.80 & 74.49 & 71.35 & 69.36 & 67.30 \\
			MGAN \cite{Fan2018} & 81.25 & 71.94 & 75.39 & 72.47 & 72.54 & 70.81 \\
			AEN \cite{Song2019} & 80.98 & 72.14 & 73.51 & 69.04 & 72.83 & 69.81 \\
			Coattention-Memnet \cite{Yang2019} & 79.70 & - & 72.90 & - & 70.50 & - \\
			DCN-CA \cite{Zhu2024} & 83.96 & 76.84 & 77.85 & 73.65 & 75.48 & 74.98 \\
			CDT \cite{Sun2019} & 82.30 & 74.02 & 77.19 & 72.99 & 74.66 & 73.66 \\
			ASGCN-DT \cite{Zhang2019} & 80.86 & 72.19 & 74.14 & 69.24 & 71.53 & 69.68 \\
			DGEDT \cite{Tang2020} & 83.90 & 75.10 & 76.80 & 72.30 & 74.80 & 73.40 \\
			Sentic-GCN \cite{Liang2022} & 84.03 & 75.38 & 77.90 & 74.71 & - & - \\
			EK-GCN \cite{Gu2023a} & 83.96 & 74.93 & 78.46 & 76.54 & 75.84 & 74.57 \\
			DGGCN \cite{Liu2023} & 83.66 & 76.73 & 75.70 & 72.57 & 74.87 & 72.27 \\
			IA-HiNET \cite{Gu2023b}  & 83.58 & 75.85 & 78.24 & 74.54 & 75.79 & 74.61 \\
			APSCL \cite{Li2023} & 83.37 & 77.31 & 77.14 & 73.86 & - & - \\\hline
			Mamba4ABSA & 82.11 & 74.72 & 76.98 & 73.11 & 74.12 & 73.50 \\
			MambaFormer & 82.93 & 75.33 & 77.53 & 73.42 & 74.47 & 73.86 \\
			\textbf{MambaForGCN (ours)} & \textbf{84.38} & \textbf{77.47} & \textbf{78.64} & \textbf{76.61} & \textbf{75.96} & \textbf{74.77} \\\hline
			BERT \cite{Devlin2018} & 85.79 & 80.09 & 79.91 & 76.00 & 75.92 & 75.18 \\
			KDGN+BERT \cite{Wu2023} & 85.79 & 80.09 & 79.91 & 76.00 & 75.92 & 75.18 \\
			EK-GCN+BERT \cite{Gu2023a} & 87.01 & 81.94 & 81.32 & 77.59 & 77.64 & 75.55 \\
			DGGCN+BERT \cite{Liu2023} & 87.65 & 82.55 & 81.30 & 79.19 & 75.89 & 75.16 \\
			DCN-CA+BERT \cite{Zhu2024} & 86.89 & 80.32 & 81.50 & 78.51 & 76.94 & 75.07 \\
			IA-HiNET+BERT \cite{Gu2023b} & \textbf{87.72} & \textbf{82.65} & 81.53 & 77.97 & 77.59 & 76.85 \\
			APSCL+BERT \cite{Li2023} & 86.79 & 81.84 & 79.45 & 76.56 & 75.88 & 75.36 \\\hline
			\textbf{MambaForGCN+BERT} & 86.68 & 80.86 & \textbf{81.80} & \textbf{78.59} & \textbf{77.67} & \textbf{76.88} \\
			\hline
		\end{tabular}
	}
	\caption{Experimental results comparison on three publicly available datasets}
	\label{tab:experimental_results}
\end{table*}

Additional 30-dimensional embeddings for position and part-of-speech (POS) are concatenated and fed into a BiLSTM model with a hidden size of 50, applying a dropout rate of 0.7 to reduce overfitting. The architecture includes SynGCN and MambaFormer Module, each with 2 layers and dropout of 0.1 and 0.05 (MHA with 4 heads). The Mamba layer features 2 convolutional filters and a 16-dimensional state vector. Model weights are uniformly initialized, and the model is trained using the Adam optimizer (Kingma \& Ba, 2014) with a 0.002 learning rate and a batch size of 16 over 50 epochs. For MambaForGCN+BERT, BERT extracts word representations from the last hidden states. Simplified versions include Mamba4ABSA (removing MHA and SynGCN) and MambaFormer (removing SynGCN). The implementation is done using PyTorch.

\subsection{Experimental Results}

Table \ref{tab:experimental_results} displays the comparison's findings with each baseline model. The accuracy and macro-averaged F1 score serve as the primary evaluation criteria for the ABSA models. First, MambaForGCN significantly improves sentiment classification accuracy compared to the syntax-based models DGEDT, Sentic-GCN, DGGCN, and the semantic-based model DCN-CA. This suggests that MambaForGCN's Mamba and MHA blocks help better capture short and long-range dependencies between aspect and opinion word relationships. Second, the gain in accuracy on three datasets indicates that the KAN-gated fusion effectively filters noise and promotes information flow between SynGCN and MambaFormer modules in ABSA. Lastly, we can see that the fundamental BERT has outperformed specific ABSA models by a considerable margin. When our MambaForGCN is combined with BERT, the outcomes demonstrate that this model's efficacy is further enhanced.

\subsection{Ablation Study}

We performed ablation experiments on the datasets to examine the effects of various components in our MambaForGCN model on performance, as shown in Table \ref{tab:ablation_study}. The phrase "w/o MHA" describes how the MHA block in the MambaFormer module has been removed. This entails using the representation from the mamba block and SynGCN module. Similarly, "w/o Mamba" involves excluding the Mamba block from the MambaFormer module, thereby using MHA and SynGCN module. Additionally, "w/o gated fusion" indicates using a fully connected network to integrate representations from the two modules without employing the KAN fusion gate. The results are shown in Table \ref{tab:ablation_study}. Notably, without MHA, the performance of MambaForGCN experiences a decrease of 1.59\%, 1.43\%, and 1.13\% for the Restaurant, Laptop, and Twitter datasets, respectively. Furthermore, the MHA layer’s representation in the MambaFormer module must be integrated with the mamba layer’s representation in the MambaFormer module, as the performance of MambaForGCN decreases by 1.71\%, 1.68\%, and 1.41\%, respectively, when solely relying on MHA in MambaFormer module. Finally, MambaForGCN performance drops by 1.90\%, 1.58\%, and 1.81\% when a primary, fully connected network is substituted for the gated fusion module. Overall, MambaForGCN performs better in capturing short and long-range dependencies between aspect and opinion words for ABSA when all components are effectively combined. It can adaptively integrate two features (syntax and semantics) from the GCN and MambaFormer.

\begin{table}[H]
	\centering
	\renewcommand{\arraystretch}{1.2} 
	\resizebox{\columnwidth}{!}{ 
		\begin{tabular}{@{}lccc@{}} 
			\hline
			\textbf{Model} & \textbf{Rest14 Acc.} & \textbf{Lapt14 Acc.} & \textbf{Twit Acc.} \\\hline
			MambaForGCN & 84.38 & 78.64 & 75.96 \\
			w/o MHA & 82.79 & 77.21 & 74.83 \\
			w/o Mamba & 82.67 & 76.96 & 74.55 \\
			w/o KAN gated fusion & 82.48 & 77.06 & 74.15 \\\hline
		\end{tabular}
	}
	\caption{Results of an ablation study (\%)}
	\label{tab:ablation_study}
\end{table}

\subsection{Effect of MambaForGCN Layer}

\begin{figure}[t]
  \includegraphics[width=\columnwidth]{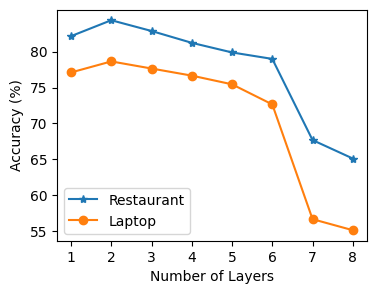}
  \caption{Effect of different numbers of MambaForGCN layers}
  \label{fig:mambaforgcn_layers}
\end{figure}

In our investigation, as depicted in Fig. \ref{fig:mambaforgcn_layers}, we observe that the Laptop and Restaurant datasets produced the best results with two layers. When the number of layers is too low, dependency information won't be adequately communicated. When the number of layers in the model is too high, it becomes overfit, and redundant information passes through, which lowers performance. Many trials must be carried out to determine an appropriate layer number.

\begin{table*}[h!]
    \centering
    \resizebox{\textwidth}{!}{%
        \begin{tabular}{p{9cm} c c c}
            \hline
            \textbf{Text} & \textbf{W/O Mamba} & \textbf{Our Model} & \textbf{Labels} \\\hline
            I trust the \textcolor{green}{[people]}\textsubscript{pos} at Go Shushi, it never disappoints. & (Pos\checkmark) & (Pos\checkmark) & (Pos) \\\hline
            Just scribbled 27 sides of pure bullshit in a two and a half hour exam, my right arm looks like one of \textcolor{red}{[Madonna]}\textsubscript{neg}'s. & (Neu$\times$) & (Neg\checkmark) & (Neg) \\\hline
            The two \textcolor{red}{[waitress]}\textsubscript{neg}'s looked like they had been sucking on lemons. & (Neg\checkmark) & (Neg\checkmark) & (Neg) \\\hline
            Great \textcolor{green}{[performance]}\textsubscript{pos} and quality. & (Pos\checkmark) & (Pos\checkmark) & (Pos) \\\hline
            I Hollywood prefers miss goody two shoes to bad girls: Now bad girls like Tara Reid, \textcolor{red}{[Paris Hilton]}\textsubscript{neg}, Britney & (Neu$\times$) & (Neg\checkmark) & (Neg) \\\hline
        \end{tabular}%
    }
    \caption{Case studies of our MambaForGCN model and ablated MambaForGCN without the Mamba module.}
    \label{tab:case_study}
\end{table*}

\subsection{Case Study}
To evaluate the efficacy of MambaForGCN in capturing long-range dependencies between aspects and opinion words enhancing ABSA, we conducted a case study using a few sample sentences. Table 4 presents the predictions and corresponding truth labels for these sentences. In the second sample, "Just scribbled 27 sides of pure bullshit in a two and a half hour exam, my right arm looks like one of Madonna’s," the aspect is "Madonna." The opinion is implied rather than explicitly stated, but it can be inferred that it relates to your arm's physical state or appearance. The sentence structure separates the aspect (Madonna) from the context that describes the opinion (your arm looking muscular or overworked). The actual descriptive comparison (implied opinion) depends on understanding the cultural reference to Madonna's muscular arms, which comes after a fair amount of text. So, there is a long-range dependency between the aspect "Madonna" and the implied opinion of "muscular" or "overworked," even though the specific opinion words aren't directly adjacent to the aspect in the sentence. Capturing this relationship requires handling long-range dependency between the aspect and the implied opinion words.
MambaForGCN adeptly determines the polarity of the aspect word “Madonna” and the opinion words by integrating the Mamba module, which successfully captured the long-range dependency. In contrast, MambaForGCN, without the Mamba module, failed to determine the polarity of the aspect “Madonna”.

\section{Conclusion}
This paper proposes the MambaForGCN framework, which integrates syntactic structure and semantic information for the ABSA tasks. We utilize SynGCN to enrich the model with syntactic knowledge. Then, we merge the selective space model (Mamba) and transformer to extract semantic information from the input and capture short and long-range dependencies. Furthermore, we fuse these modules with a KAN-gated feature fusion to maximize their interaction and filter out irrelevant information. The outcomes of our experiments show that our method works well on three publicly available datasets.

\section*{Limitation}
The drawback of this work is the potential difficulty in generalization to diverse real-world datasets. Although MambaForGCN demonstrates effectiveness on three benchmark datasets, its performance may vary when applied to texts with different linguistic patterns, domain-specific terminologies, or out-of-vocabulary words not covered in the training data.

\section*{Acknowledgments}
The authors thank the anonymous reviewers for their helpful comments. A big gratitude to Qiaolan Meng. This work is supported by the National Science Foundation of China (62177002).


\end{document}